# Large Deviation Methods for Approximate Probabilistic Inference

Michael Kearns and Lawrence Saul
{mkearns,lsaul}@research.att.com
AT&T Labs — Research
180 Park Avenue
Florham Park, NJ 07932

## Abstract

We study two-layer belief networks of binary random variables in which the conditional probabilities $\Pr[\text{child}|\text{parents}]$ depend monotonically on weighted sums of the parents. In large networks where exact probabilistic inference is intractable, we show how to compute upper and lower bounds on many probabilities of interest. In particular, using methods from large deviation theory, we derive rigorous bounds on marginal probabilities such as $\Pr[\text{children}]$ and prove rates of convergence for the accuracy of our bounds as a function of network size. Our results apply to networks with generic transfer function parameterizations of the conditional probability tables, such as sigmoid and noisy-OR. They also explicitly illustrate the types of averaging behavior that can simplify the problem of inference in large networks.

## 1 Introduction

The intractability of probabilistic inference in general graphical models [2] has led to several interesting lines of research examining specialized algorithms for restricted classes of models. One such line of work has the well-known polytree algorithm [9] for exact inference as its starting point, and can be viewed as studying the limiting case of *sparse* networks — models that either start with relatively low connectivity, or can be massaged through graph-theoretic operations into equivalent models with low connectivity [6]. Another, rather different, approach has been the study of the limiting case of *dense* networks. Such networks arise naturally in a variety of contexts, including medical diagnosis [11] and the modeling of cortical spike train data [3]. Inspired by ideas from statistical mechanics and convex duality, several so-called *variational* methods have been introduced for dense networks [4, 5, 10], along with guarantees that they provide rigorous upper and lower bounds on the desired probabilities of interest.

The current work is a contribution to this latter line of results on dense networks. We study two-layer [1] belief networks of binary random variables in which the conditional probabilities $\Pr[\text{child}|\text{parents}]$ depend monotonically on weighted sums of the parents. In large networks, where exact probabilistic inference is intractable, we show how to compute upper and lower bounds on various probabilities of interest. This is done by exploiting the averaging phenomena that occur at nodes with many parents. Since our bounds rely on the introduction of auxiliary parameters associated with each local conditional probability table, our approach can be viewed as a particular type of variational method for approximate probabilistic inference. The approach we take in this paper, however, has a number of distinguishing features: for example, it applies quite generally to the computation of both upper and lower bounds on marginal probabilities, as well as to generic transfer function parameterizations of the conditional probability tables (including sigmoid and noisy-OR). We also prove quantitative *rates of convergence* of the accuracy of our bounds as a function of network size. Finally, and perhaps most important from a conceptual standpoint, is that our approach explicitly illustrates (and exploits) the types of "averaging" behavior that can occur in large belief networks.

Our results are derived by applying the theory of large deviations — generalizations of well-known tools such as Hoeffding and Chernoff bounds [1] — to the weighted sums of parents at each node in the network. At each node with parents, we introduce a variational parameter that essentially quantifies what it means for the incoming weighted sum to fall "near" its mean.

---

[1] We have also generalized all of our results to the multi-layer case. These generalizations will be presented elsewhere.

Bounds on the marginal probabilities are then computed from two contributions: one assuming that all of the weighted sums throughout the network fall near their mean values, and the other assuming that they do not. These contributions give rise to an interesting trade-off between probable explanations of the evidence and improbable deviations from the mean. In particular, in networks where each child has $N$ parents, the gap between our upper and lower bounds behaves as a sum of two terms, one of order $\sqrt{\gamma \ln(N)/N}$, and the other of order $N^{1-2\gamma}$, where $\gamma$ is a free parameter. The choice $\gamma = 1$ yields a $O(\sqrt{\ln(N)/N})$ convergence rate; but more generally, all of the variational parameters are chosen to optimize the previously mentioned trade-off.

In addition to providing such rates of convergence for *large* networks, our methods also suggest efficient algorithms for approximate inference in *fixed* networks.

The outline of the paper is as follows: in Section 2, we give standard definitions for two-layer belief networks with parametric conditional probability tables. In Section 3, we derive the large-deviation results we will use. In Section 4, we present our main results: inference algorithms for two-layer networks, along with proofs that they compute lower and upper bounds on marginal probabilities, and derivations of rates of convergence. In Section 5, we give a brief experimental demonstration of our ideas. In Section 6, we conclude with a discussion and some remarks on the relationship between our work and previous variational methods.

## 2 Definitions and Preliminaries

A belief network is a graph-theoretic representation of a probabilistic model, in which the nodes represent random variables, and the links represent causal dependencies. The joint distribution of this model is obtained by composing the local conditional probability distributions, **Pr**[child|parents], specified at each node in the network. In principle, in networks where each node represents a discrete random variable, these conditional probability distributions can be stored as table of numbers whose rows sum to one. In practice, however, representing the distributions in this way can be prohibitively expensive.

For networks of binary random variables, so-called *transfer functions* provide a convenient way to parameterize large conditional probability tables.

**Definition 1** *A* **transfer function** *is a mapping* $f : [-\infty, \infty] \to [0, 1]$ *that is everywhere differentiable and satisfies* $f'(x) \geq 0$ *for all $x$ (thus, $f$ is nondecreasing). If $f'(x) \leq \alpha$ for all $x$, we say that $f$ has* **slope** $\alpha$.

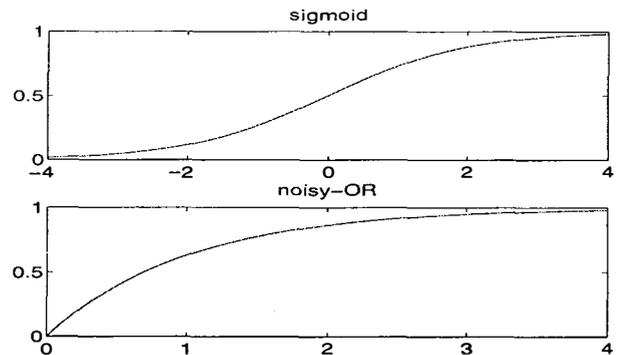

Figure 1: Plots of the sigmoid and noisy-OR transfer functions.

Common transfer functions used in belief networks include the noisy-OR transfer function, $f(x) = 1 - e^{-x}$, and the sigmoid $f(x) = 1/(1 + e^{-x})$ [7]; plots of these function are shown in Figure 1.

Because the value of a transfer function is bounded between 0 and 1, it can be interpreted as the conditional probability that a binary random variable takes on a particular value. In particular, in networks of binary random variables, transfer functions can be used to parameterize conditional probability tables in which **Pr**[child|parents] depends monotonically on a weighted sum of the parents. This leads us to consider the following class of two-layer belief networks:

**Definition 2** *For any transfer function $f$, a* **two-layer probabilistic $f$-network** *is defined by:*

- *Binary* **input** *variables $X_1, \ldots, X_N$ and* **output** *variables $Y_1, \ldots, Y_M$.*

- *For every input $X_j$ and output $Y_i$, a real-valued* **weight** $\theta_{ij}$ *from $X_j$ to $Y_i$.*

- *For every input $X_j$, a* **bias** $p_j$.

*A two-layer probabilistic $f$-network $C$ defines a joint probability distribution over all of the variables $\{X_j\}$ and $\{Y_i\}$ as follows: each input variable $X_j$ is independently set to 1 with probability $p_j$, and to 0 with probability $1 - p_j$. Then given binary values $X_j \in \{0, 1\}$ for all of the inputs, the output $Y_i$ is set to 1 with probability $f(\sum_{j=1}^{N} \theta_{ij} X_j)$.*

In general, we shall use $j$ as an index over inputs, and $i$ as an index over outputs. An example of a two-layer, fully connected network is given in Figure 2. In a noisy-OR network, the weight $\theta_{ij} = -\ln \mathbf{Pr}[Y_i = 1 | X_j = 1, X_{j' \neq j} = 0]$ represents the probability that the $i$th output is set to one given that only its $j$th input is set to one. In a sigmoid network, the weight $\theta_{ij}$ represents the $j$th parameter in a logistic regression for the output $Y_i$.

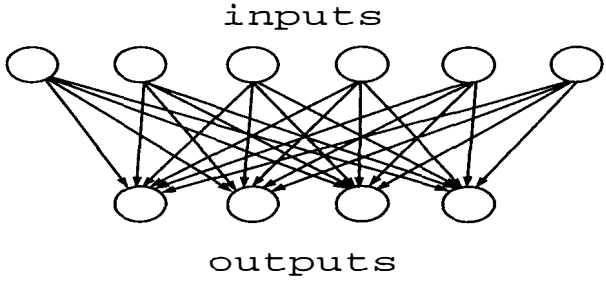

Figure 2: A two-layer, fully-connected belief network.

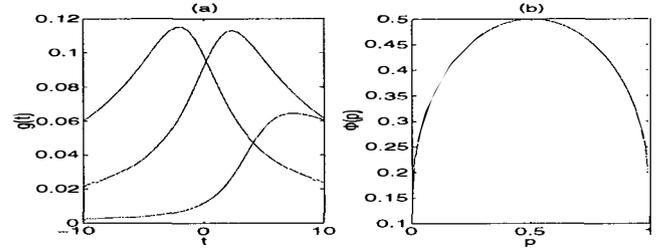

Figure 3: Plots of (a) $g(t) = t^{-2} \ln[(1-p)e^{-pt} + pe^{(1-p)t}]$ for different fixed values of $p$; (b) $\Phi(p) = (1-2p)/\ln\left(\frac{1-p}{p}\right)$.

Note that the weighted sums of inputs, $\sum_{j=1}^{N} \theta_{ij} X_j$, play a special role in belief networks with transfer function conditional probability tables; in particular, we have the independence relation:

$$\mathbf{Pr}\left[Y_i | \sum_{j=1}^{N} \theta_{ij} X_j, X_1, \ldots, X_N\right] = \mathbf{Pr}\left[Y_i | \sum_{j=1}^{N} \theta_{ij} X_j\right]. \quad (1)$$

As we shall see, many useful results can be derived simply by using this independence relation in combination with the convergence properties of weighted sums of binary random variables. This directly motivates the subject of the next section.

## 3 Large Deviation Bounds

In this section, we derive large deviation results for weighted sums of binary random variables. The following preliminary lemma will prove extremely useful.

**Lemma 1** *For all $p \in [0,1]$ and $|t| < \infty$,*

$$\frac{1}{t^2} \ln\left[(1-p)e^{-tp} + pe^{t(1-p)}\right] \leq \frac{1}{4}\Phi(p) \quad (2)$$

*where $\Phi(p) = (1-2p)/\ln((1-p)/p)$.*

**Proof:** Let $g(t)$ denote the left hand side of Equation (2). Figure 3(a) shows some plots of $g(t)$ versus $t$ for different (fixed) values of $p$. The maximum of $g(t)$ is determined by the vanishing of the derivative: $g'(t) = 0$. It is straightforward to show that this occurs at $t^* = 2\ln\left(\frac{1-p}{p}\right)$. Evaluating $g(t^*)$ gives the desired result. A plot of $\Phi(p)$ is shown in Figure 3(b). □

Equipped with this lemma, we can now give a simple upper bound on the probability of large deviations for weighted sums of independent binary random variables. The following theorem generalizes classical large-deviation results [1], and will serve as the starting point for our analysis of two-layer networks in Section 4.

**Theorem 2** *For all $1 \leq j \leq N$, let $X_j \in \{0,1\}$ denote independent binary random variables with means $p_j$, and let $|\theta_j| < \infty$. Then for all $\epsilon > 0$:*

$$\mathbf{Pr}\left[\left|\frac{1}{N}\sum_{j=1}^{N} \theta_j(X_j - p_j)\right| > \epsilon\right] \leq 2e^{-N\epsilon^2/\chi^2} \quad (3)$$

*where $\chi^2 = \frac{1}{N}\sum_{j=1}^{N} \theta_j^2 \Phi(p_j)$.*

**Proof:** Consider first the probability that $\frac{1}{N}\sum_{j=1}^{N} \theta_j(X_j - p_j) > \epsilon$. Note that for any random variable $X$, we have $\mathbf{Pr}[X > 0] = \frac{1}{2}\mathbf{E}[1 + X/|X|]$, where $\mathbf{E}[\cdot]$ denotes the expectation. Now let $\eta > 0$ be any positive number. Noting that $\frac{1}{2}(1+X/|X|) \leq e^{\eta X}$, we have:

$$\mathbf{Pr}\left[\frac{1}{N}\sum_{j=1}^{N} \theta_j(X_j - p_j) > \epsilon\right]$$

$$\leq \mathbf{E}\left[e^{\eta \sum_{j=1}^{N} \theta_j(X_j - p_j) - \eta N \epsilon}\right] \quad (4)$$

$$= e^{-\eta N \epsilon} \mathbf{E}\left[\prod_{j=1}^{N} e^{\eta \theta_j(X_j - p_j)}\right] \quad (5)$$

$$= e^{-\eta N \epsilon} \prod_{j=1}^{N} \mathbf{E}\left[e^{\eta \theta_j(X_j - p_j)}\right] \quad (6)$$

$$= e^{-\eta N \epsilon} \prod_{j=1}^{N} \left[(1-p_j)e^{-\eta \theta_j p_j} + p_j e^{\eta \theta_j(1-p_j)}\right] \quad (7)$$

$$\leq e^{-\eta N \epsilon + \frac{\eta^2}{4}\sum_{j=1}^{N} \theta_j^2 \Phi(p_j)}. \quad (8)$$

Equations (5) — (7) follow from the properties of exponentials and the assumption that $X_j$ are independently distributed with means $p_j$. The last inequality follows from applying Lemma 1 to each of the terms in Equation (7). Note that this last inequality holds for any positive value of $\eta$; in particular, it holds for $\eta = 2\epsilon/\chi^2$, where $\chi^2 = \frac{1}{N}\sum_{j=1}^{N} \theta_j^2 \Phi(p_j)$. Substituting this value of $\eta$ into the inequality gives the bound

$$\mathbf{Pr}\left[\frac{1}{N}\sum_{j=1}^{N} \theta_j(X_j - p_j) > \epsilon\right] \leq e^{-N\epsilon^2/\chi^2}. \quad (9)$$

Finally, we note that using similar methods, one can derive an identical upper bound on the probability of negative $\epsilon$-deviations. Adding the two together gives the desired result. □

The large deviation bound in Equation (3) states that $\epsilon$-deviations from the mean become exponentially improbable as $e^{-N\epsilon^2/\chi^2}$. The parameter $\chi^2$ summarizes the dependence of the bound on the weights $\theta_j$ and the probabilities $p_j$. Note that $\chi^2$ vanishes when the weighted sum $\sum_{j=1}^{N} \theta_j X_j$ has zero variance. This happens when—for all $j$—either the weight $\theta_j$ is equal to zero, or the probability $p_j$ is equal to zero or one. In this case, even for finite $N$, there is zero probability of $\epsilon$-deviations from the mean, and the bound in Equation (3) capture this explicitly. More generally, when the weights $\theta_j$ are of order unity and the probabilities $p_j$ are bounded away from zero and one, then $\chi^2$ is of order unity. In particular, in the uniform case where $\theta_j = 1$ and (say) $p_j = 1/2$, one recovers the standard Chernoff bounds for sums of i.i.d. random variables.

## 4   Bounds on Marginal Probabilities

In this section, we apply the large deviation results from Section 3 to make approximate probabilistic inferences in two-layer belief networks with transfer function conditional probability tables. In particular, we derive rigorous upper and lower bounds on marginal probabilities of evidence at the output layer, and also prove quantitative rates of convergence for these bounds to the true probability.

Consider the generic two-layer network shown in Figure 2. Appealing to a medical analogy, in such a network, a typical "diagnostic" query might be to assess the likelihood of a certain "disease" $X_j$ based on the observed "symptoms" $Y = \{Y_i\}$. This amounts to computing the posterior probability:

$$\mathbf{Pr}[X_j|Y] = \frac{\mathbf{Pr}[X_j]\mathbf{Pr}[Y|X_j]}{\mathbf{Pr}[Y]}. \quad (10)$$

In principle, the factors on the right hand side can be computed by summing over possible instantiations of the unlabeled inputs; for example,

$$\mathbf{Pr}[Y] = \sum_X \mathbf{Pr}[X,Y] \quad (11)$$

where here $X = \{X_j\}$. Generally, though, the amount of computation required to perform this sum scales exponentially with the number of inputs, $N$. The same intractability applies to the probability $\mathbf{Pr}[Y|X_j = 1]$ in the numerator of Equation (10). Thus we are naturally motivated to consider approximate schemes for probabilistic inference in these networks.

In what follows, we state all our results in terms of upper and lower bounds on marginal probabilities of evidence observed at the output layer. We consider the case where evidence is instantiated at a subset of $K \leq M$ outputs, which without loss of generality we will assume are $Y_1, \ldots, Y_K$. Though we state our results in terms of the marginal probabilities $\mathbf{Pr}[Y_1, Y_2, \ldots, Y_K]$, it is worth emphasizing that our methods apply equally well to the calculation of *conditional* marginal probabilities, such as $\mathbf{Pr}[Y_1, Y_2, \ldots, Y_K | X_j]$. (The latter can simply be viewed as an unconditional marginal probability in a two-layer belief network with $N-1$ unobserved inputs.) Thus, by propagating our bounds on these marginal probabilities through Bayes rule (Equation (10)), we can compute upper and lower bounds on posterior probabilities that are expressed as ratios of marginal probabilities. Thus our framework for approximate probabilistic inference can support queries that are either diagnostic ($\mathbf{Pr}[X_i|Y]$) or predictive ($\mathbf{Pr}[Y|X_i]$) in nature. One important qualification, however, is that the rates of convergence we derive apply only to marginal probabilities, and not posterior probabilities.

It will be most convenient to express our results by introducing the scaling $\theta_{ij} = \tau_{ij}/N$, where $\tau_{ij} \in [-\tau_{max}, \tau_{max}]$. This simply expresses the weights as order $1/N$ quantities, rendering the weighted sum of inputs to any output an order 1 quantity bounded in absolute value by the parameter $\tau_{max}$.

Let us briefly describe, at a high level, our approach for upper and lower bounding the marginal probability. We will argue that, over the random independent draws of the values at the input layer, with all but a (hopefully small) "throw-away" probability, the weighted sums of inputs entering *every* output unit are "close" to their expected values. Now *conditioned* on these weighted sums being near their means, we can easily compute worst-case upper and lower bounds on the marginal probability. The throw-away probability is so named because conditioned on some weighted sum falling far from its mean, we will simply lower bound the marginal probability by 0 and upper bound it by 1. Our overall bounds are obtained by taking the appropriate weighted combination of the bounds for these mutually exclusive cases. There is a nontrivial and interesting competition between the two terms.

We divide the exposition into three parts. First, in Section 4.1, we derive *families* of both upper and lower bounds on the marginal probability of the evidence. These families are parameterized by the choice of a parameter $\epsilon_i$ for each evidence variable $Y_i$. Different choices of the $\epsilon_i$ lead to different bounds.

The parametric families of upper and lower bounds provided in Section 4.1 immediately suggest compu-tationally efficient algorithms for computing upper and lower bounds on the evidence probability for a *given* choice of the $\epsilon_i$, but do not immediately indicate choices for these parameters leading to *good* upper and lower bounds. Thus in Section 4.2, we examine the bounds that result from some specific, easily computed (but perhaps non-optimal) choices for the $\epsilon_i$. We prove that these choices lead to upper and lower bounds on the evidence probability whose *difference* can be bounded by nice functions of natural properties of the network, such as its size. For example, for any $\gamma > 1$, we are able to make choices for the $\epsilon_i$ leading to upper and lower bounds that differ at most by $2\alpha\tau_{max}\beta^{K-1}K\sqrt{\gamma\ln(N)/N} + 2K/N^{2\gamma}$, where $N$ and $K$ are as above, and $\beta < 1$. These are the first results giving bounds on the *rate* of convergence of the gap between rigorous upper and lower bounds as a function of network size, and they suggest the possibility of certain "universal" behavior for such convergence, such as a leading dependence on $N$ that is an inverse square root power law. This behavior is reminiscent of the large body of literature on learning curves in supervised learning, which is not surprising, since our underlying weapon is large deviation bounds.

As desirable as it is to have specific choices of $\epsilon$ that lead to upper and lower bounds that converge rapidly with network size, in practice the network size is fixed, and these bounds may be weak. Thus, in Section 4.3 we propose efficient algorithms based on our parametric bounds that are designed to compute the tightest possible bounds in a fixed network. The main idea is to simply perform gradient descent/ascent on the parameters $\epsilon_i$ to optimize the upper/lower bounds.

### 4.1 A Parametric Family of Bounds

We begin by stating our most general upper and lower bounds on marginal probabilities. As we have discussed, we will actually give a *family* of bounds, parameterized by a choice of values $\epsilon_1, \ldots, \epsilon_K > 0$; we think of $\epsilon_i$ as a parameter associated with the output $Y_i$. After stating our bounds in this general form, we will gradually show their power by making specific, simple choices for the parameters $\epsilon_i$ that give quantitative "rates" of convergence to the true marginal probabilities that depend on the network size.

**Theorem 3** *Let $f$ be a transfer function, and let $C$ be a 2-layer probabilistic $f$-network with $N$ inputs, $M$ outputs, bias $p_j$ on input $X_j$, and weight $\theta_{ij} = \tau_{ij}/N$ from input $X_j$ to output $Y_i$, where $\tau_{ij} \in [-\tau_{max}, \tau_{max}]$. Then for any subset $\{Y_1, \ldots, Y_K\}$ of the outputs of $C$, any setting $v_1, \ldots, v_K$, and any $\epsilon_1, \ldots, \epsilon_K > 0$, the marginal probability $\mathbf{Pr}[Y_1 = v_1, \ldots, Y_K = v_K]$ obeys*

$$\mathbf{Pr}[Y_1 = v_1, \ldots, Y_K = v_K]$$

$$\leq \left(1 - 2\sum_{i=1}^{K} e^{-N\epsilon_i^2/\chi_i^2}\right) \times$$

$$\prod_{v_i=1} f(\mu_i + \epsilon_i) \prod_{v_i=0} (1 - f(\mu_i - \epsilon_i))$$

$$+ 2\sum_{i=1}^{K} e^{-N\epsilon_i^2/\chi_i^2} \qquad (12)$$

*and*

$$\mathbf{Pr}[Y_1 = v_1, \ldots, Y_K = v_K]$$

$$\geq \left(1 - 2\sum_{i=1}^{K} e^{-N\epsilon_i^2/\chi_i^2}\right) \times$$

$$\prod_{v_i=1} f(\mu_i - \epsilon_i) \prod_{v_i=0} (1 - f(\mu_i + \epsilon_i)) \quad (13)$$

*where $\mu_i = \sum_{j=1}^{N} \theta_{ij} p_j$ and $\chi_i^2 = \frac{1}{N}\sum_{j=1}^{N} \tau_{ij}^2 \Phi(p_j)$.*

We give the proof of this theorem below, but first discuss its form and implications briefly. Let us use $P_C^U(\epsilon_1, \ldots, \epsilon_K)$ to denote the upper bound on the marginal probability $\mathbf{Pr}[Y_1 = v_1, \ldots, Y_K = v_K]$ given by the right-hand side of Equation (12), and let us use $P_C^L(\epsilon_1, \ldots, \epsilon_K)$ to denote the lower bound on the marginal probability given by the right-hand side of Equation (13).

First, note that these equations immediately give very efficient algorithms for computing $P_C^U$ and $P_C^L$ for given choices of the $\epsilon_i$. Second, note that there is no reason to use the same set of $\epsilon_i$ in both $P_C^U$ and $P_C^L$ — that is, the optimal choices for these parameters (yielding the bounds closest to the true probability) may differ in the upper and lower bounds. We will not exploit this fact when we make simple choices for the parameters leading to specific rates of convergence, but it will be an important issue when discussing algorithms for fixed networks in Section 4.3. Third, both $P_C^U$ and $P_C^L$ involve nontrivial trade-offs in the $\epsilon_i$. For example, in $P_C^L$, the factor $(1 - 2\sum_{i=1}^{K} e^{-N\epsilon_i^2/\chi_i^2})$ is clearly maximized by choosing all $\epsilon_i$ as large as possible, while the remaining products of transfer functions are maximized by choosing all $\epsilon_i = 0$. Similar trade-offs appear in $P_C^U$.

We now give the proof of Theorem 3.

**Proof:** Consider the probability that over a random draw of the inputs $X_j$, one or more of the weighted sums $\sum_{j=1}^{N} \theta_{ij} X_j$ lies further than $\epsilon_i$ away from its mean value $\mu_i = \sum_{j=1}^{N} \theta_{ij} p_j$. We can upper bound this probability as follows:

$$\mathbf{Pr}\left[\left|\sum_{j=1}^{N} \theta_{ij}(X_j - p_j)\right| > \epsilon_i \text{ for some } 1 \leq i \leq N\right]$$

$$\leq \sum_{i=1}^{K} \mathbf{Pr}\left[\left|\sum_{j=1}^{N} \theta_{ij}(X_j - p_j)\right| > \epsilon_i\right] \quad (14)$$

$$\leq 2\sum_{i=1}^{K} e^{-N\epsilon_i^2/\chi_i^2}. \quad (15)$$

In Equation (14), we have use the so-called *union bound*, which simply states that the probability of a union of events is bounded by the sum of their individual probabilities. The next line follows directly from the result of Theorem 2.

Suppose that for all $1 \leq i \leq K$, the weighted sums $\sum_{j=1}^{N} \theta_{ij} X_j$ lie in the intervals $[\mu_i - \epsilon_i, \mu_i + \epsilon_i]$. Conditioned on this event, the probability of evidence $Y_1 = v_1, \ldots, Y_K = v_K$ is at most:

$$\prod_{v_i=1} f(\mu_i + \epsilon_i) \prod_{v_i=0} [1 - f(\mu_i - \epsilon_i)] \quad (16)$$

This is because the transfer function $f$ is non-decreasing, so subject to the constraint that the weighted sum of inputs to every output is within $\epsilon_i$ of its mean, we maximize the probability of the evidence by taking this weighted sum to be $\epsilon_i$ above the mean for positive findings, and $\epsilon_i$ below the mean for negative findings. Similarly, we can lower bound the probability of the evidence subject to this constraint by reversing the orientations of the $\epsilon_i$.

Now suppose that one or more of the weighted sums $\sum_{j=1}^{N} \theta_{ij} X_j$ lies outside the intervals $[\mu_i - \epsilon_i, \mu_i + \epsilon_i]$. In this event (which occurs with probability at most $2\sum_{i=1}^{K} e^{-N\epsilon_i^2/\chi_i^2}$), we can clearly upper bound the probability of the evidence by 1, and lower bound it by 0.

To derive bounds on the overall marginal probability, we simply combine the bounds for these two different cases—the one in which the weighted sums fall within their $\epsilon$-intervals, and the other in which they do not. We clearly obtain upper and lower bounds by assuming that the latter case occurs with the maximum probability given by the union bound. This gives rise to the weighted combinations in Equations (12) and (13). □

The proof of Theorem 3 provides a very simple "two-point" intuition for our upper and lower bounds. For each output $Y_i$, we are approximating the *distribution of its weighted sum of inputs* by just two points: one point $\epsilon_i$ above or below (depending on whether we are deriving the upper or lower bound) the mean $\mu_i$, and the other point at either $-\infty$ or $+\infty$. The relative weight of these two points in our approximation of the true distribution of the weighted sum depends on the choice of $\epsilon_i$: as $\epsilon_i$ becomes smaller, we give more weight to the $\pm\infty$ point, with the trade-off governed by the union bound sum. We can think of the weight given to the $\pm\infty$ point as a "throw-away" probability, since we always upper or lower bound the marginal probability by 1 or 0 here.

It is possible to generalize our methods to provide $r$-point approximations to the marginal probabilities. The resulting inference algorithms scale exponentially with $r$, but provide tighter bounds. As an aside to the specialist, we observe that while our methods *approximate* the integral of the *true* transfer function at a small number of points, some previous variational methods compute the *exact* integral of an *approximation* to the true transfer function [5].

Theorem 3 provides our most general parametric bounds, and will form our starting point in our discussion of algorithms for fixed networks in Section 4.3. For now, however, we would like to move towards proposing algorithms making specific choices of the $\epsilon_i$ that yield useful quantitative bounds on the difference $P_C^U - P_C^L$. For this purpose, it will be useful to present the following parametric family of bounds on this difference that exploits bounded slope transfer functions.

**Theorem 4** . *Let $f$ be a transfer function of slope $\alpha$, and let $C$ be a two-layer probabilistic $f$-network defined by $\{\theta_{ij}\}$ and $\{p_j\}$. Then for any $\epsilon_1, \ldots, \epsilon_K$,*

$$P_C^U(\epsilon_1, \ldots, \epsilon_K) - P_C^L(\epsilon_1, \ldots, \epsilon_K)$$

$$\leq 2\alpha \sum_{i=1}^{K} \left( \epsilon_i \prod_{v_j=1, j \neq i} f(\mu_j + \epsilon_j) \times \right.$$

$$\left. \prod_{v_j=0, j \neq i} [1 - f(\mu_j - \epsilon_j)] \right)$$

$$+ 2\sum_{i=1}^{K} e^{-N\epsilon_i^2/\chi_i^2} \quad (17)$$

*where $\mu_i = \sum_{j=1}^{N} \theta_{ij} p_j$ and $\chi_i^2 = \frac{1}{N}\sum_{j=1}^{N} \tau_{ij}^2 \Phi(p_j)$.*

Let us again briefly discuss the form of this bound on $P_C^U - P_C^L$ before giving its proof. The first term in Equation (17) is essentially twice the slope times a sum of the $\epsilon_i$. Actually, it is considerably better than this, since each $\epsilon_i$ is multiplied by a product of $K$ factors smaller than 1 (the $f(\mu_j + \epsilon_j)$ and $[1 - f(\mu_j - \epsilon_j)]$). The second term is the union bound sum. There is again a clear competition between the sums: the first is minimized by choosing all $\epsilon_i$ as small as possible, while the second is minimized by choosing all $\epsilon_i$ as large as possible.

Equation (17) also permits another important observation. For $K$ fixed but as $N \to \infty$, we may clearly allow $\epsilon_i \to 0$ for all $i$, and in this limit we have $P_C^U - P_C^L \to 0$. In Section 4.2, we will have more to say about the *rate* of approach. Here we simply observe that in this limit, the upper and lower bounds are both approaching the product of the factors $f(\mu_j)$ and $[1 - f(\mu_j)]$ — in other words, our results demonstrate that in the limit of large $N$, the output distribution simply *factorizes*. The simplicity of this limit accounts for the strength of our bounds for large but finite $N$.

We now give the proof of Theorem 4.

**Proof:** The second term of Equation (17) is simply the union bound sum, which appears in $P_C^U(\epsilon_1, \ldots, \epsilon_K)$ but not $P_C^L(\epsilon_1, \ldots, \epsilon_K)$. The remaining difference is

$$\left(1 - 2\sum_{i=1}^{K} e^{-N\epsilon_i^2/\chi_i^2}\right) \times$$
$$\left(\prod_{v_i=1} f(\mu_i + \epsilon_i) \prod_{v_i=0} [1 - f(\mu_i - \epsilon_i)]\right.$$
$$\left. - \prod_{v_i=1} f(\mu_i - \epsilon_i) \prod_{v_i=0} [1 - f(\mu_i + \epsilon_i)]\right) \quad (18)$$

We will simply bound the factor $(1 - 2\sum_{i=1}^{K} e^{-N\epsilon_i^2/\chi_i^2})$ by 1, and concentrate on the remaining difference of products. To do this, we will start with the smaller product (where for convenience and without loss of generality we will assume that the positive findings are on the outputs $Y_1, \ldots, Y_{K'}$, and the negative findings are on $Y_{K'+1}, \ldots, Y_K$)

$$\prod_{i=1}^{K'} f(\mu_i - \epsilon_i) \prod_{i=K'+1}^{K} [1 - f(\mu_i + \epsilon_i)] \quad (19)$$

and "walk" $K$ steps towards the larger product, bounding the increase at each step. Thus, consider the intermediate product

$$\prod_{i=1}^{R} f(\mu_i + \epsilon_i) \prod_{i=R+1}^{K'} f(\mu_i - \epsilon_i) \times$$
$$\prod_{i=K'+1}^{S} [-f(\mu_i - \epsilon_i)] \prod_{i=S+1}^{K} [1 - f(\mu_i + \epsilon_i)] (20)$$

Here we have already walked towards the larger product on the positive findings $Y_1, \ldots, Y_R$ and on the negative findings $Y_{K'+1}, \ldots, Y_S$. Now consider the single additional step of changing the factor $f(\mu_{R+1} - \epsilon_{R+1})$ in the smaller product to its desired form in the larger product, $f(\mu_{R+1} + \epsilon_{R+1})$. This change results in a product that differs from that given in Equation (20) by exactly

$$(f(\mu_{R+1} + \epsilon_{R+1}) - f(\mu_{R+1} - \epsilon_{R+1})) \times$$
$$\prod_{i=1}^{R} f(\mu_i + \epsilon_i) \prod_{i=R+2}^{K'} f(\mu_i - \epsilon_i) \times$$

$$\prod_{i=K'+1}^{S} [1 - f(\mu_i - \epsilon_i)] \prod_{i=S+1}^{K} [1 - f(\mu_i + \epsilon_i)]$$
$$\leq 2\alpha\epsilon_{R+1} \prod_{i=1,\ldots,K',i\neq R+1} f(\mu_i + \epsilon_i) \times$$
$$\prod_{i=K'+1}^{K} [1 - f(\mu_i - \epsilon_i)] \quad (21)$$

Here we have used

$$f(\mu_{R+1} + \epsilon_{R+1}) - f(\mu_{R+1} - \epsilon_{R+1}) \leq 2\alpha\epsilon_{R+1} \quad (22)$$

since $f$ has slope $\alpha$, and we have upper bounded each factor $f(\mu_i - \epsilon_i)$ by $f(\mu_i + \epsilon_i)$, and each factor $[1 - f(\mu_i + \epsilon_i)]$ by $[1 - f(\mu_i - \epsilon_i)]$. Bounding the one-step change induced by altering a factor corresponding to a negative finding is entirely analogous, and the overall bound of Equation (17) is simply the sum of the bounds on the one-step changes. □

### 4.2 Rates of Convergence

In this section, we propose a specific, easily computed choice for the parameters $\epsilon_i$ in Theorem 4, resulting in algorithms for efficiently computing $P_C^U$ and $P_C^L$ with guaranteed bounds on the gap $P_C^U - P_C^L$.

Consider the choice $\epsilon_i = \sqrt{2\gamma\chi_i^2 \ln(N)/N}$, for some $\gamma > 1$. Plugging into Equation (17) yields

$$P_C^U(\epsilon_1, \ldots, \epsilon_K) - P_C^L(\epsilon_1, \ldots, \epsilon_K)$$
$$\leq \frac{2K}{N^{2\gamma}} + 2\alpha\sqrt{\frac{2\gamma \ln N}{N}} \times$$
$$\sum_{i=1}^{K} \left(\chi_i \prod_{\substack{v_j = 1 \\ j \neq i}} f(\mu_j + \epsilon_j) \prod_{\substack{v_j = 0 \\ j \neq i}} [1 - f(\mu_j - \epsilon_j)]\right) \quad (23)$$

The second term of this bound essentially has a $1/\sqrt{N}$ dependence on $N$, but is multiplied by a damping factor that we might typically expect to decay exponentially with the number $K$ of outputs examined. To see this, simply notice that each of the factors $f(\mu_j + \epsilon_j)$ and $[1 - f(\mu_j - \epsilon_j)]$ is bounded by 1; furthermore, since all the means $\mu_j$ are bounded, if $N$ is large compared to $\gamma$ then the $\epsilon_i$ are small, and each of these factors is in fact bounded by some value $\beta < 1$. Thus the second term in Equation (23) is bounded by $2\alpha\tau_{max}\beta^{K-1}K\sqrt{2\gamma\ln(N)/N}$. Since it is natural to expect the marginal probability of interest itself to decrease exponentially with $K$, this is desirable and natural behavior.

Of course, in the case of large $K$, the behavior of the resulting overall bound can be dominated by the first term $2K/N^{2\gamma}$ of Equation (23), which does not enjoy the same exponentially decreasing damping factor. In such situations, however, we can consider larger values of $\gamma$, possibly even of order $K$; indeed, for sufficiently large $\gamma$, the second term (which scales like $\sqrt{\gamma}$) must necessarily overtake the first one. Thus there is a clear trade-off between the two terms, as well as optimal value of $\gamma$ that sets them to be (roughly) the same magnitude. Generally speaking, for fixed $K$ and large $N$, we observe a leading dependence on $N$ that takes the form of an inverse square root power law.

Of course, different parameters and architectures require different choices of the $\epsilon_i$ to get the best bounds, and we have only scratched the surface so far. In particular, we have crudely summarized all of the parameters by the bounds $\tau_{max}$ and $\beta$, and in the process ignored many of the dependencies between the $Y_i$ (and therefore, between the best choices for the $\epsilon_i$). The algorithms given by the gradient computations in the next section will account for these dependencies more strongly.

### 4.3 An Algorithm for Fixed Networks

As we discussed at the outset of this section, the specific and simple choices we made for the parameters $\epsilon_i$ in Section 4.2 in order to obtain nice theoretical bounds on the gap $P_C^U - P_C^L$ may be far from the optimal choices for a fixed network of interest. However, Theorem 3 directly suggests a natural algorithm for inference in fixed networks. In particular, regarding $P_C^U(\epsilon_1, \ldots, \epsilon_K)$ as a function of the $\epsilon_i$ determined by the network parameters, we may perform a gradient descent on $P_C^U(\epsilon_1, \ldots, \epsilon_K)$ in order to find a local minimum of our upper bound. The components $\partial P_C^U/\partial \epsilon_i$ of the gradient $\nabla P_C^U$ are easily computable for all the commonly studied transfer functions, and we must simply obey the mild and standard constraints $\epsilon_i > 0$ during the minimization.

Similar comments apply to finding a local *maximum* of the lower bound $P_C^L$, where we instead would perform a gradient *ascent*. Again the gradient is easily computed. As we have already mentioned, it is important to note that the values of the $\epsilon_i$ achieving a (local) minimum in $P_C^U$ may be quite different from the values achieving a (local) maximum in $P_C^U$.

As an example of the suggested computations, consider the lower bound $P_C^L$ given by Equation (13). For the gradient computation, it is helpful to instead maximize $\ln P_C^L$, where for a positive finding $v_j = 1$ we obtain

$$\frac{\partial \ln P_C^L}{\partial \epsilon_j} = \frac{(4N\epsilon_j/\chi_j^2)e^{-N\epsilon_j^2/\chi_j^2}}{(1 - 2\sum_{i=1}^{K} e^{-N\epsilon_i^2/\chi_i^2})} - \frac{f'(\mu_j - \epsilon_j)}{f(\mu_j - \epsilon_j)}. \quad (24)$$

The gradient components for negative findings $v_j =$

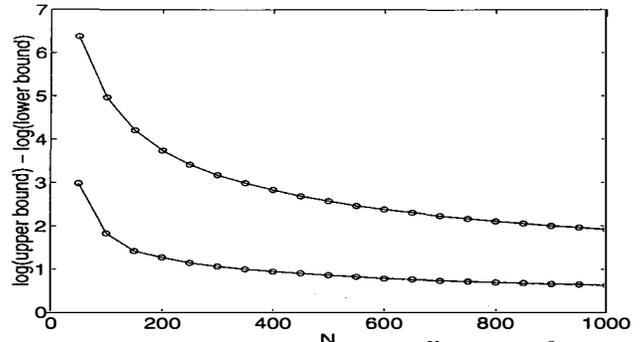

Figure 4: Plots of the difference $\log(P_C^U) - \log(P_C^L)$ as a function of $N$. Upper curve: $P_C^U$ and $P_C^L$ were obtained using the fixed choices for the $\epsilon_i$ described in Section 4.2. Upper curve: $P_C^U$ and $P_C^L$ were obtained by optimizing the $\epsilon_i$. Each point represents an average of 25 trials, as described in the text.

0 are quite similar. The gradients for the upper bound $P_C^U$ are more complex, but still easily computable. Equation (24) also highlights the fact that although our bounds are efficiently computable, they still express complex dependencies between the variables through the parameters $\epsilon_i$, as we see that the gradient with respect to $\epsilon_j$ depends on all the other $\epsilon_i$.

## 5    An Experimental Illustration

As a simple and brief experimental illustration of some of the ideas presented here, we offer the results of a simulation. In this simulation, we randomly generated two-layer sigmoidal networks with $M = 25$ outputs and a varying number of inputs $N$. The unscaled weights $\tau_{ij}$ were chosen randomly from a normal distribution with zero mean and unit variance, and thus are consistent with the scaling assumed by our formal results. For each value of $N$, 25 such random networks were generated, and for each such network, a random $M$-bit output vector was chosen as the evidence. $N$ ranged from 50 to 1000 in increments of 50.

For each network and evidence query, we computed four bounds: the upper and lower bounds derived from choosing the $\epsilon_i$ as specified in Section 4.2, and upper and lower bounds derived by optimizing Equations (12) and (13), as discussed in Section 4.3. The differences between the (natural) logarithms of these pairs of bounds, averaged over 25 random networks for each $N$, are plotted in Figure 4. As predicted by our formal results, the gaps between upper and lower bounds are diminishing nicely with increasing $N$. Note that for the largest value $N = 1000$, the best bounds are performing fairly well: while the typical finding has a probability on the order of only $2^{-M}$ (where $M = 25$), our best upper and lower bounds differ by only a factor of about 2.

# 6  Discussion

Our methods for bounding marginal probabilities rely on the introduction of auxiliary parameters — namely, the $\epsilon_i$ — that are associated with each node in the network. Both our lower and upper bounds on marginal probabilities are expressed in terms of these parameters. For very large networks, we have seen that informed (but non-optimal) choices of these parameters give rise to polynomial rates of convergence. More generally, though, for networks of fixed size, we have proposed algorithms for finding the $\epsilon_i$ that yield the tightest possible bounds on marginal probabilities.

Like previous work on variational methods[5], our approach relates the problem of probabilistic inference to one of optimization. Indeed, it is interesting (though perhaps not too surprising) that large deviation methods lead to algorithms of the same general nature as those derived from convex duality[4] and statistical mechanics[10]. Thus, for example, our $\epsilon_i$ play a similar role to the dual variables introduced by Legendre transformations of the log-transfer function. In both cases, the introduction of auxiliary or variational parameters makes it possible to perform an otherwise intractable average over hidden variables. Moreover, the correlations between hidden variables, induced by the evidence, are reflected by the coupling of variational parameters in the expressions for lower and upper bounds on marginal probabilities. Indeed, the balance we have observed between probable explanations of the evidence and improbably large deviations resembles previously encountered trade-offs, such as the competition between energy and entropy in statistical mechanical approaches.

Having noted the similarities of our approach to previous work, it is worth pointing out certain differences. Our methods are novel in several respects. First, they apply in essentially the same way to the calculation of both lower and upper bounds on marginal probabilities. Moreover, they apply quite generally to monotonic transfer functions; we neither require nor exploit any further properties such as log-concavity. Second, the large deviation methods come with rates of convergence and performance guarantees as a function of network size. In some sense, our results provide a formal justification for earlier claims, based mainly on intuition, that variational methods can work well in large directed graphical models.

Of course, large deviation methods also have their own limitations. They are designed mainly for very large probabilistic networks whose weights are of order $1/N$, where $N$ is the number of parents at each node in the graph. The limit of large $N$ considered in this paper is analogous to the thermodynamic limit for physical (undirected graphical) models of infinite-range ferromagnets[8]. In both directed and undirected graphical models, weights of order $1/N$ give rise to the simplest type of limiting behavior as $N \to \infty$. It should be noted, however, that other limits (for instance, weights of order $1/\sqrt{N}$) are also possible. The main virtue of the large deviation methods is that they explicitly illustrate the types of averaging behavior that occur in certain densely connected networks. This suggests that such networks, though hopelessly intractable for exact probabilistic inference, can serve as useful models of uncertainty.

# References

[1] Cover, T., & Thomas J. (1991) *Elements of Information Theory*. New York: John Wiley.

[2] Dagum, P., & Luby, M. (1993) Approximating probabilistic inference in Bayesian belief networks is NP-hard. *Artificial Intelligence, 60*, 141–153.

[3] De Sa, V., DeCharms, R.C., & Merzenich. M. (1998) Using Helmholtz machines to analyze multi-channel neuronal recordings. To appear in M.I. Jordan, M. J. Kearns, & S.A. Solla, Eds., *Advances in Neural Information Processing Systems 10*, MIT Press.

[4] Jaakkola, T. (1997) *Variational methods for inference and estimation in graphical models*. Unpublished doctoral dissertation, MIT.

[5] Jordan, M., Ghahramani, Z., Jaakkola, T., & Saul, L. (1997) An introduction to variational methods for graphical models. To appear in M. Jordan, ed. *Learning in Graphical Models*, Kluwer Academic.

[6] Jensen, F.V. (1996) *An Introduction to Bayesian Networks*. London: UCL Press.

[7] Neal, R. (1992) Connectionist learning of belief networks. *Artificial Intelligence, 56*, 71–113.

[8] Parisi, G. (1988) *Statistical field theory*. Redwood City, CA: Addison-Wesley.

[9] Pearl, J. (1988) *Probabilistic Reasoning in Intelligent Systems: Networks of Plausible Inference*. San Mateo, CA: Morgan Kaufmann.

[10] Saul, L., Jordan, M., and Jaakkola, T. (1996) Mean field theory for sigmoid belief networks. *Journal of Artificial Intelligence Research 4*, 61–76.

[11] Shwe, M.A., Middleton, B., Heckerman, D.E., Henrion, M., Horvitz, E.J., Lehmann, H.P., & Cooper, G.F. (1991) Probabilistic diagnosis using a reformulation of the INTERNIST-1/QMR knowledge base. I. The Probabilistic Model and Inference Algorithms. *Methods in Information and Medicine, 30*, 241–255.